\documentclass[10pt, a4paper]{article}
\usepackage{IOA}
\usepackage{blindtext}
\usepackage{layout}
\usepackage{xcolor}
\usepackage{graphicx}
\usepackage[space]{grffile}
\usepackage{float}
\usepackage{blindtext}
\usepackage{placeins}
\usepackage{amsmath}								
\usepackage{amssymb}	
\usepackage[style=numeric,backend=bibtex,sorting=none]{biblatex}
\usepackage{authblk}							

\usepackage{xcolor}

\title{Measuring Human Assessed Complexity in Synthetic Aperture Sonar Imagery Using the Elo Rating System}

\firstauthor{BT Reinhardt}
\secondauthor{ID Gerg}
\thirdauthor{DC Brown}
\fourthauthor{JD Park}
\fifthauthor{}
\sixthauthor{}

\firstaffl{Applied Research Labs Penn State, State College PA, USA}
\secondaffl{Applied Research Labs Penn State, State College PA, USA}
\thirdaffl{Applied Research Labs Penn State, State College PA, USA}
\fourthaffl{Applied Research Labs Penn State, State College PA, USA}
\fifthaffl{}
\sixthaffl{}

\begin{document}

	\maketitle 
	\makeauthors

\section{Introduction}

Performance of automatic target recognition from synthetic aperture sonar data is heavily dependent on the complexity of the beamformed imagery \cite{Stack:2011a,Midtgaard:2014a}. Operators describe the task of explaining image complexity in the context of minehunting as a difficult problem.  Several mechanisms can contribute to this including: unwanted vehicle dynamics, complex acoustic channels, the presence of biological activity, the bathymetry of the scene, and the presence of natural and manmade clutter.  One often has an intuitive sense of what makes a particular image difficult, but determining a reliable mathematical formula for the task has proven elusive.

In order to understand the impact of the environmental complexity on automatic image perception, researchers have taken approaches rooted in information theory \cite{donderi2006information, keogh2004towards, Myers:2007a} and heuristics \cite{Peters:1990a}. Despite these efforts, a quantitative measure for complexity has not been related to the phenomenology from which it is derived.  By using subject matter experts (SMEs) we will derive a complexity metric for a set of imagery which is associated to the perceived complexity. The goal of this work is to develop an understanding of how several common information theoretic and heuristic measures are related to the SME perceived complexity in synthetic aperture sonar imagery.

We use the following approach to develop an SME-based complexity metric.  First, an ensemble of 10 meter x 10 meter images were cropped from a SAS data set that spans multiple environments. Second, we choose to have operators compare pairs of images from a finite set and have them rank the images in one of three categories with respect to minehunting: image A is more complex than image B, image B is more complex than image A, or images A and B have the same complexity.  We do this because it is difficult for humans to rank order large sets of information (Miller's Law) \cite{miller1956magical}.  Many methods exist to estimate the rank order of a set of pair-wise comparisons.  One such method, Elo ranking, was originally developed for rank ordering chess players based on the outcome of matches (win, loose, or draw) \cite{Elo:1978a}.  Thus, finally, we translate these comparisons into a complexity metric using Elo ranking.

The Elo method produced a plausible rank ordering across the broad dataset, despite some disagreement between operators on which conditions lead to the most complex environment (e.g. coral versus sand ripples).  Heuristic and information theoretical metrics were then compared to the Elo score metric. The metrics with highest degree of correlation were those relating to spatial information, e.g. variations in pixel intensity, with an R-squared value of approximately 0.8. However, this agreement was dependent on the scale from which the spatial variation was measured, with only optimal kernel sizes being demonstrated here. We will also be presenting a comparison between the Elo score and many other metrics including lacunarity, image compression, and entropy.

\section{Methodology} 

\subsection{Human Assessment of Complexity} 

A web application was created which presents a human with two SAS images and asks which image is more complex: the image on the left, right, or are both images similar in complexity. The images were dynamic range compressed (DRC) by normalizing to the interval $[0,1]$ and then applying the following
\begin{equation}
\textrm{drc\_pixel\_value} = 20 \cdot \log_{10}(\textrm{pixel\_value})+\epsilon.
\end{equation}

\begin{figure}[h]
	\begin{center}
		\includegraphics[width=0.85\textwidth]{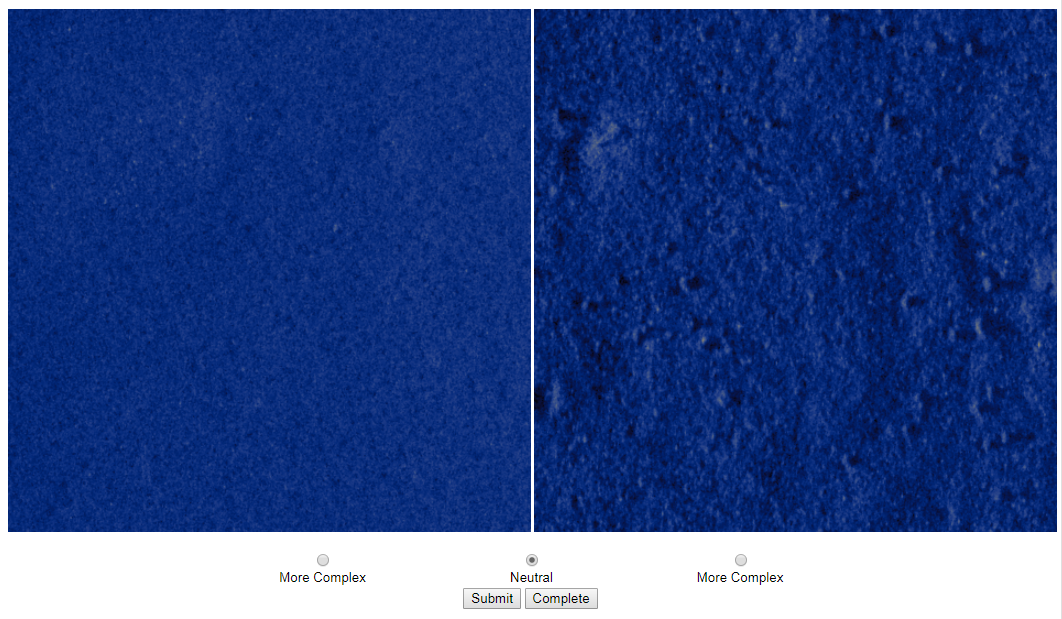}	
	\end{center}
	\caption{This image is a depiction of the SME interface used to obtain the training data for this analysis. A pair of SAS images are presented to the user, and they rate the relative complexity for minehunting.}
	\label{fig:gui1}
\end{figure}

Figure \ref{fig:gui1} shows a screenshot of the tool which presented the imagery to the human and allowed the human to select the complexity.  The SME was able to choose between three radio buttons to indicate which of the images was either more complex or neutral, neutral indicating the SME could not indicate which of the images were more complex.  The tool stored results in a text file which was later parsed into Elo scores. The tool was built using HTML and Javascript and it ran in a browser.

The tool randomly selects two images from the database to display to the SME.  The next images displayed are randomly selected from the image database with replacement.  There were a total of 6,786 possible comparisons.

\subsection{Image Elo Scores}
Rank ordering a large set of images is infeasible to carry out by a human \cite{miller1956magical}.  Instead, images are ranked utilizing a set of pairwise comparisons from a subset of all possible comparisons.  Fortunately, a number of competitive sports derive ranking systems based on matches or, in our case, pairwise comparisons. One such method is the Elo rating system first proposed by Arpad Elo in the 1960s as an improved method for ranking chess competitors \cite{Elo:1978a}. In this system, each competition between a pair of participant's results in an update to each participants rating. The participants have rating of $R_{i,n}$ and $R_{j,n}$ prior to the competition. These ratings, or Elo scores, are then updated based on the outcome of the competition using the formula
\begin{gather}
  R_{i,n+1} = R_{i,n} + K(\hat{W}_{ij}-W_{ij}) \\
  R_{j,n+1} = R_{j,n} + K(\hat{W}_{ji}-W_{ji}),
  \label{eq:EloEq}
\end{gather}
where $K$ is a scale factor to determine the overall impact of a single event. $W_{ij}$ is the expected outcome of the event which is determined by using participants scores in a logistic function:

\begin{equation}
    W_{ij} = \frac{1}{10^{-\frac{R_{i,n}-R_{j,n}}{400}}+1}.
\end{equation}
$W_{ij}$ is a function of the score difference and $\hat{W}_{ij}$ is the result of the competition. Here it is simply three states with values for win, loss, or draw based on the two team's scores $S_i$ and $S_j$
\begin{equation}
  \hat{W}_{ij} =
    \begin{cases}
        1 & \mbox{if } S_i-S_j > 0  \\
        0.5 & \mbox{if } S_i-S_j = 0  \\
        0 & \mbox{if } S_i-S_j < 0. \\
    \end{cases}
\end{equation}
A direct corollary is drawn between the competitors and the images, but instead of the Elo ranking system modeling the underlying talent of each player, it models the underlying complexity of each image.  One drawback of this ranking method is its sensitivity to the order in which images are compared or fed through the ranking process.  Whereas a player's underlying talent can evolve over time, the complexity of an image does not.  Therefore, the Elo score was computed N times for each image, with the set of comparisons being randomly shuffled for each new ranking.  The true Elo score is estimated as the mean of the distribution of Elo scores for each image.

\subsection{Dataset}
The data used in this analysis is from a high frequency synthetic aperture sonar collected across a diverse set of environments along the Pacific, Gulf of Mexico, and Atlantic coastlines of the USA.  The environments are distinguished in Table \ref{table:sites}.

\begin{table}[h]
\centering
\begin{tabular}{|c|l|}
\hline
Site Label &  Description   \\
\hline
Site A   &  Benign bottom   \\
Site B   &  Medium scale ripples   \\
Site C   &  Benign with sub-sediment structure and clutter   \\
Site D   &  Diverse environment with coral, medium ripples, and bioturbation   \\
Site E   &  Fine scale ripples   \\
\hline
\end{tabular}
\caption{This table is a list of the environments surveyed during the data collection with a brief description of their characteristics.}
\label{table:sites}
\end{table}

A total of 117 image chips were selected from the environments from Table \ref{table:sites}.  Each image chip was selected randomly from 10m to 40m in range to avoid the water column as well as other unwanted artifacts.  It was insured that images had high quality before be inserted into the dataset by manually selecting images.  All image chips with significant sensor cross talk, uncompensated motion or lack of spectral support were removed from the dataset.

Comparisons between the SMEs were made by computing the Pearson correlation coefficient between the set of overlapping comparisons made among the operators.  During testing, an SME was shown a set of images previously compared to measured self consistency.  Table \ref{table:pearsonCC} shows the computed Pearson correlation coefficients.  Subject matter experts 5 and 6 had poor self consistency and were not used for subsequent analysis.  Subject matter experts 2, 3, and 4 were all self consistent and mostly in agreement with the relative complexity of the images, for the most part.  Subject matter expert 1 was self consistent but had a different idea on what constituted a complex environment.

\begin{table}[h]
\centering
\begin{tabular}{|l|c|c|c|c|c|c|}
\hline
     & Op 1 & Op 2 & Op 3 & Op 4 & Op 5 & Op 6   \\
\hline
Op 1 & 0.86 & 0.13 & 0.50 & 0.44 & 0.79 & 0.45  \\
Op 2 &      & 1.00 & 0.72 & 0.78 & 0.58 & 0.70  \\
Op 3 &      &      & 0.79 & 0.75 & 0.25 & 0.60  \\
Op 4 &      &      &      & 0.92 & 0.47 & 0.62  \\
Op 5 &      &      &      &      & -0.1 & 0.30  \\
Op 6 &      &      &      &      &      & 0.38  \\
\hline
\end{tabular}
\caption{A table showing the Pearson correlation coefficients between SMEs comparisons.}
\label{table:pearsonCC}
\end{table}

A set of six subject matter experts compared approximately 500-1,000 images each, giving 5,722 total comparisons. Each image had received an average of 99 comparisons with a minimum of 70 comparisons.  The Elo scores are computed by Equation \ref{eq:EloEq} which is dependent on the order in which the images are compared.  Therefore, the Elo score is computed a thousand times, each time randomly changing the order in which the images are compared.  The mean score is retained as the Elo score for each image.

\section{Elo Ranking Results}

The result of the rankings seemed reasonable when spot checked. Further, the Elo scores from each of the sites seemed reasonable with the observed bottom characteristics. For instance, ripples consistently had higher Elo scores than benign bottoms.  Test site A was a relatively benign sandy bottom with patches of bioturbation.  Test site C was a benign sandy bottom with a few clutter objects while test site B had larger ripples.  Test site D had a very dynamic bottom texture including ripples, bioturbation, benign sections, and coral.  At test site E the bottom was relatively benign with fine ripples.

\begin{figure}[h]
	\begin{center}
		\begin{tabular}{c c c}
			\includegraphics[width=0.3\textwidth]{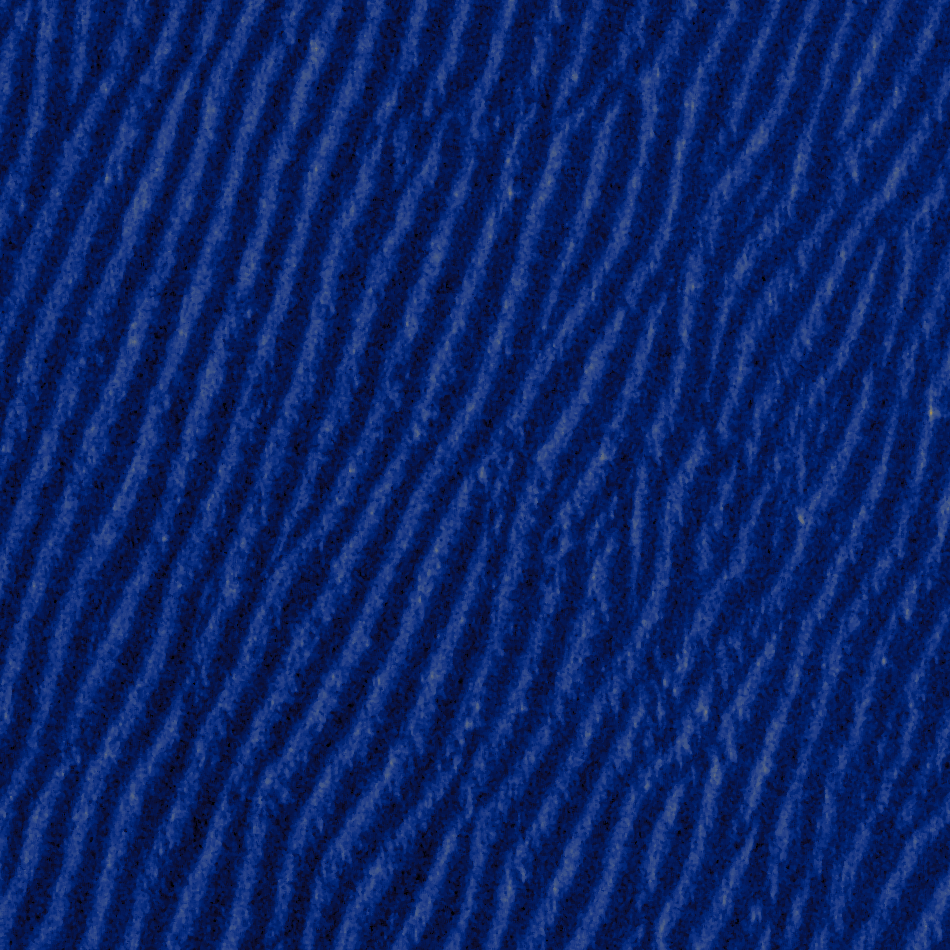}	 & \includegraphics[width=0.3\textwidth]{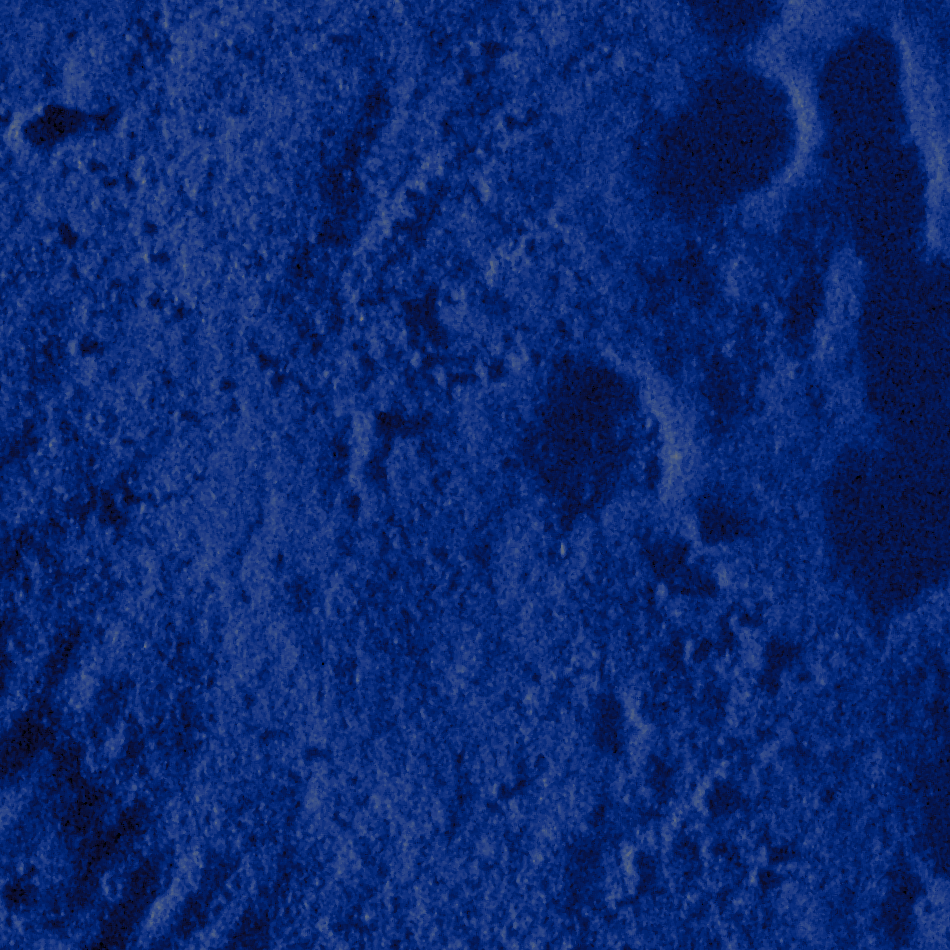}	&
			\includegraphics[width=0.3\textwidth]{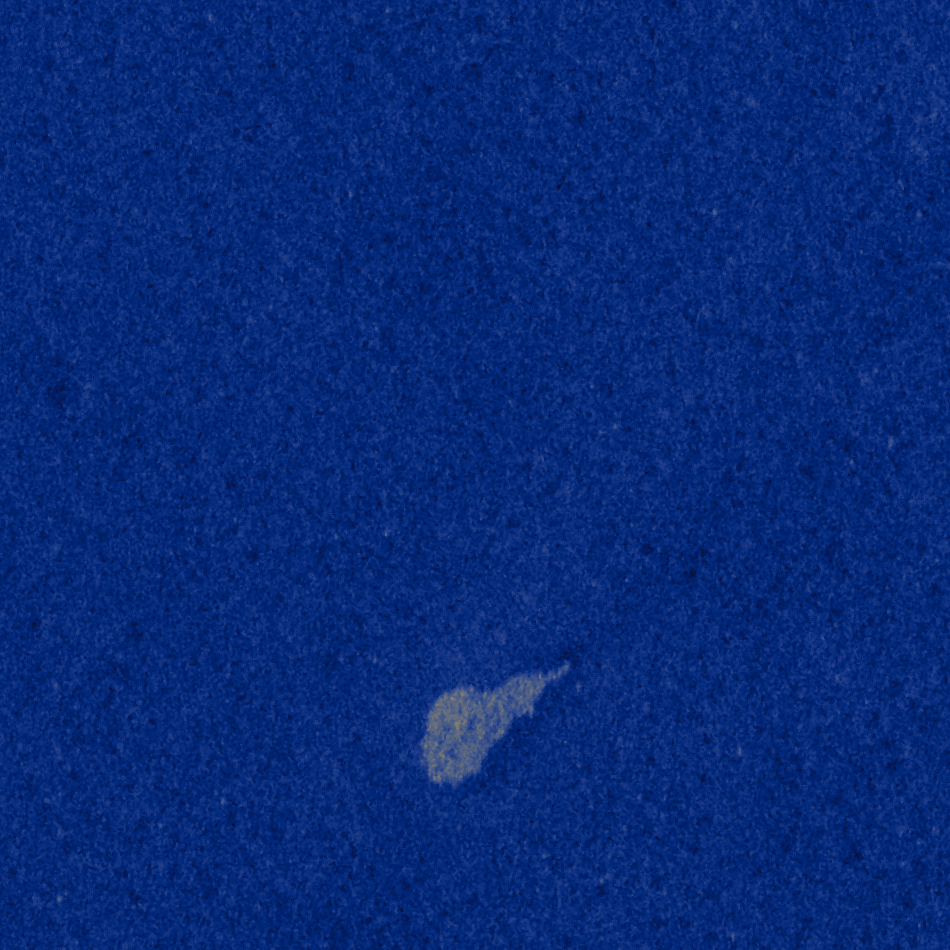}	\\
			(a) B, R = 1273 & (b) D, R = 1062 & (c) C, R = 925 \\			
			\newline & \newline & \newline \\
        \end{tabular}
        \begin{tabular}{c c}
			\includegraphics[width=0.3\textwidth]{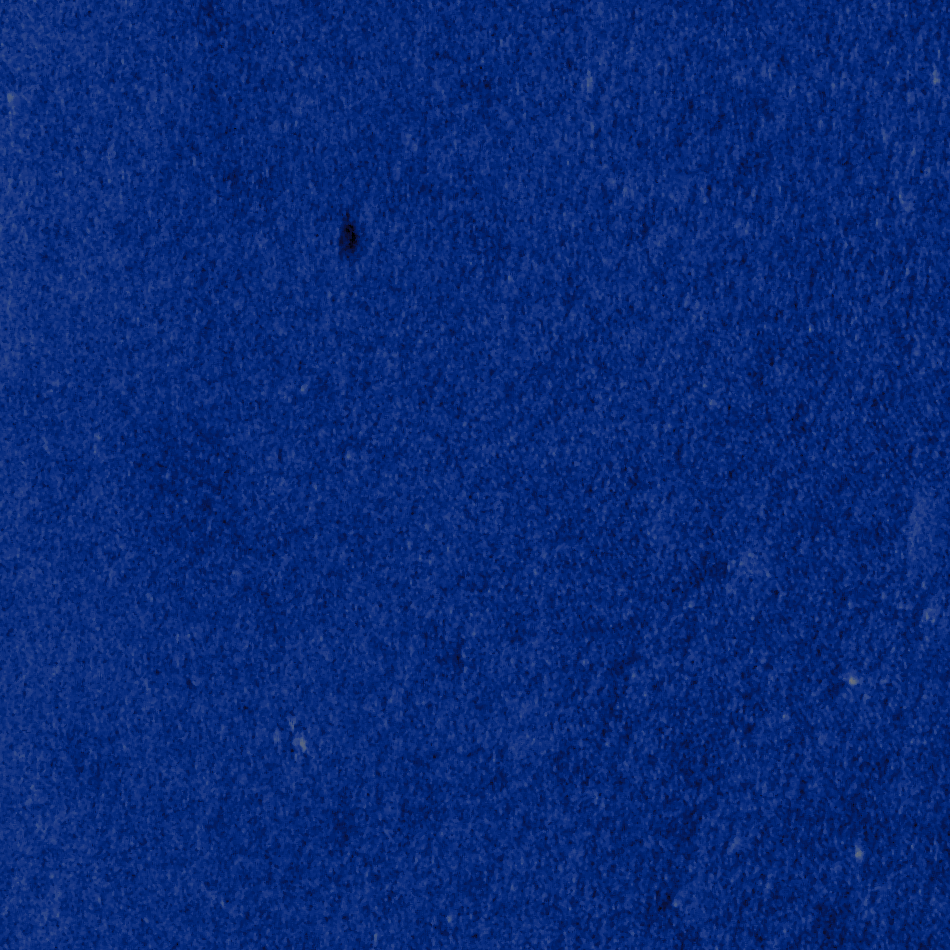}	 & \includegraphics[width=0.3\textwidth]{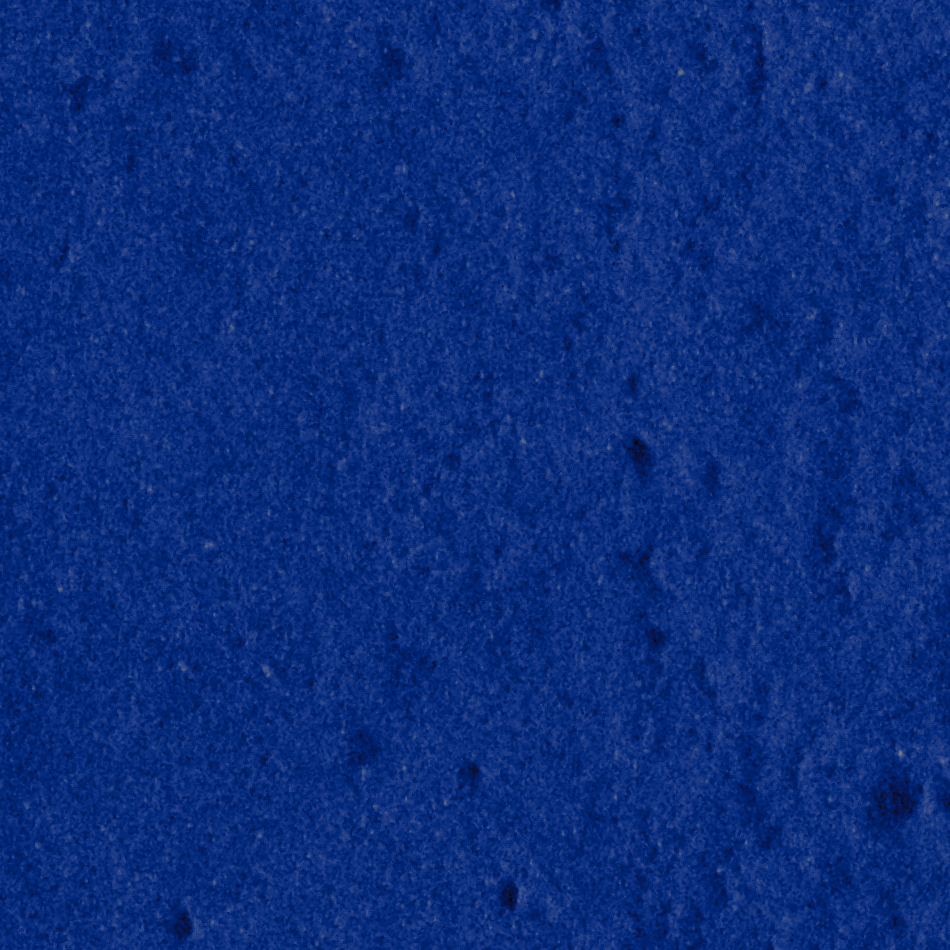} \\		
			(d) E, R = 894 & (e) A, R = 869 \\		
	    \end{tabular}		
	\end{center}
	\caption{Five image chips are shown sampled from across the range of Elo complexity scores.  The variable R is the mean Elo score and the preceding letter corresponds to the test site the image was derived from.}
	\label{fig:chipsWithScore}
\end{figure}

\begin{figure}[h]
  \centering
  \includegraphics[width=0.85\textwidth]{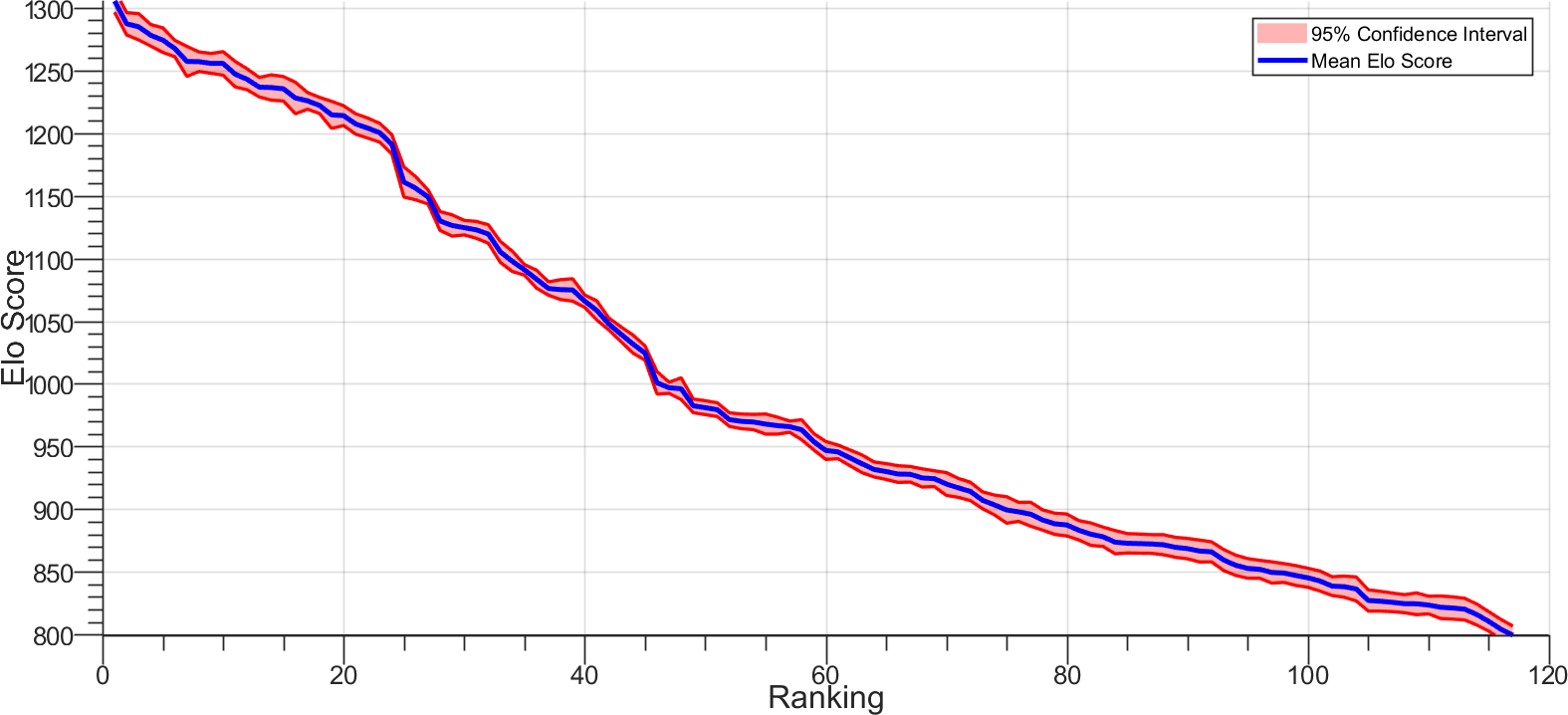}
  \caption{Elo score of images in rank order with confidence intervals.}\label{fig:eloScore}
\end{figure}

\begin{figure}[h]
  \centering
  \includegraphics[width=0.85\textwidth]{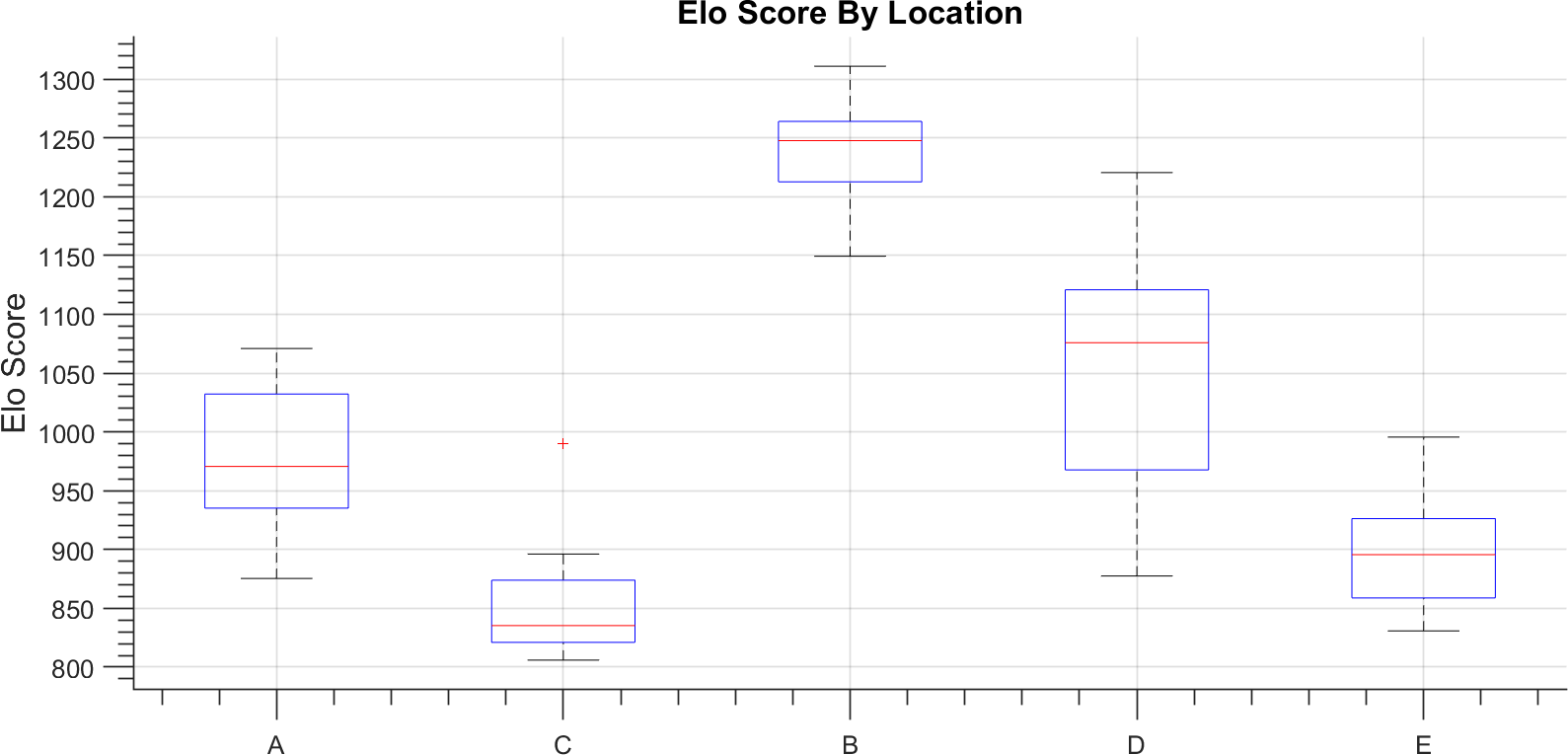}
  \caption{Elo score box plot organized by site.  The box represents the second and third quantiles while the whiskers correspond to approximately 99.3 percent of the data.}\label{fig:eloScoreSite}
\end{figure}

The Elo scores ranges from 800 to 1300 with a relatively uniform distribution.  The scores are shown in Figure \ref{fig:eloScore} with confidence intervals determined by shuffling the order in which the images are used to compute the Elo score as mentioned previously.  These scores were accumulated and sorted by site.  Figure \ref{fig:eloScoreSite} shows the relative complexity of each of the sites as determined by the Elo score.
Test site D, which had the highest diversity of bottom textures, had the largest range in computed Elo scores.  Test site B which is characterized by large ripples was determined to be the most complex by the operators, while Test site C which was a benign sand bottom had the lowest Elo Scores.

\section{Comparison of Elo Score and Common Complexity Metrics}
\subsection{Image Metrics}
Many metrics have been used to characterize image complexity \cite{Peters:1990a,Midtgaard:2014a,donderi2006information, keogh2004towards, Myers:2007a}.  Some of these include lacunarity, edge intensity, entropy and the jpeg compression ratio.  Lacunarity and edge intensity are measures of image texture. Lacunarity was popularized by Madelbrot and relates to how the pixel intensities fill the image \cite{mandelbrot1982fractal}, while, edge intensity is more related to the prominence of structure in an image.  The compression ratio and entropy are common measures used to characterize the complexity of an image that are rooted an information theoretic framework.  Both are estimates of information content of the image.

Lacunarity, $\mathcal{L}$, is a measure of the spatial structure of the intensity in an image.  In this work, it is computed using the integral image method developed by Williams \cite{Williams:2015a}.  There has been good success using this parameter to distinguishing between seabed environments for image segmentation \cite{peeples2018possibilistic}.

Edge intensity, $\mathcal{E}$, is an alternative measure for characterizing the intensity variations in an image.  It has been shown to correlate well with the complexity of an image \cite{yu2013image}. In this work, $\mathcal{E}_k$ is computed using an adjustable Sobel filter computed with a 1.5 m x 1.5 m kernel. The magnitude of the vertical and horizontal edge intensities is averaged over the 10 m by 10 m image chip.

Entropy, $\Gamma$, is a common measure of the information content.  Researchers have recently showed success in using entropy to characterize the structural complexity of natural images.  They accomplished this by taking the difference between the joint entropy and the mutual entropy between neighboring pixels normalized by the maximum distribution for the mutual and joint entropies \cite{proulx2008measures}.

Lastly, the compression ratio, $\mathcal{CR}$, is measured by first filtering the intensity image using a median filter with a 15 pixel kernel. The purpose of this filter is remove fine scaled features such as speckle. The image is then colorized and saved as a lossless PNG and a lossy JPEG. The compression ratio is defined as the ratio of the size of the lossy compressed image to the lossless compressed image. Alternatively, one can normalize the compression ratio by the root mean squared error in lossy compressed image ($\mathcal{CR}_{RMSE}$) \cite{yu2013image}.

\subsection{Regression Results}
\begin{figure}[htbp]
  \centering
  \includegraphics[width=0.85\textwidth]{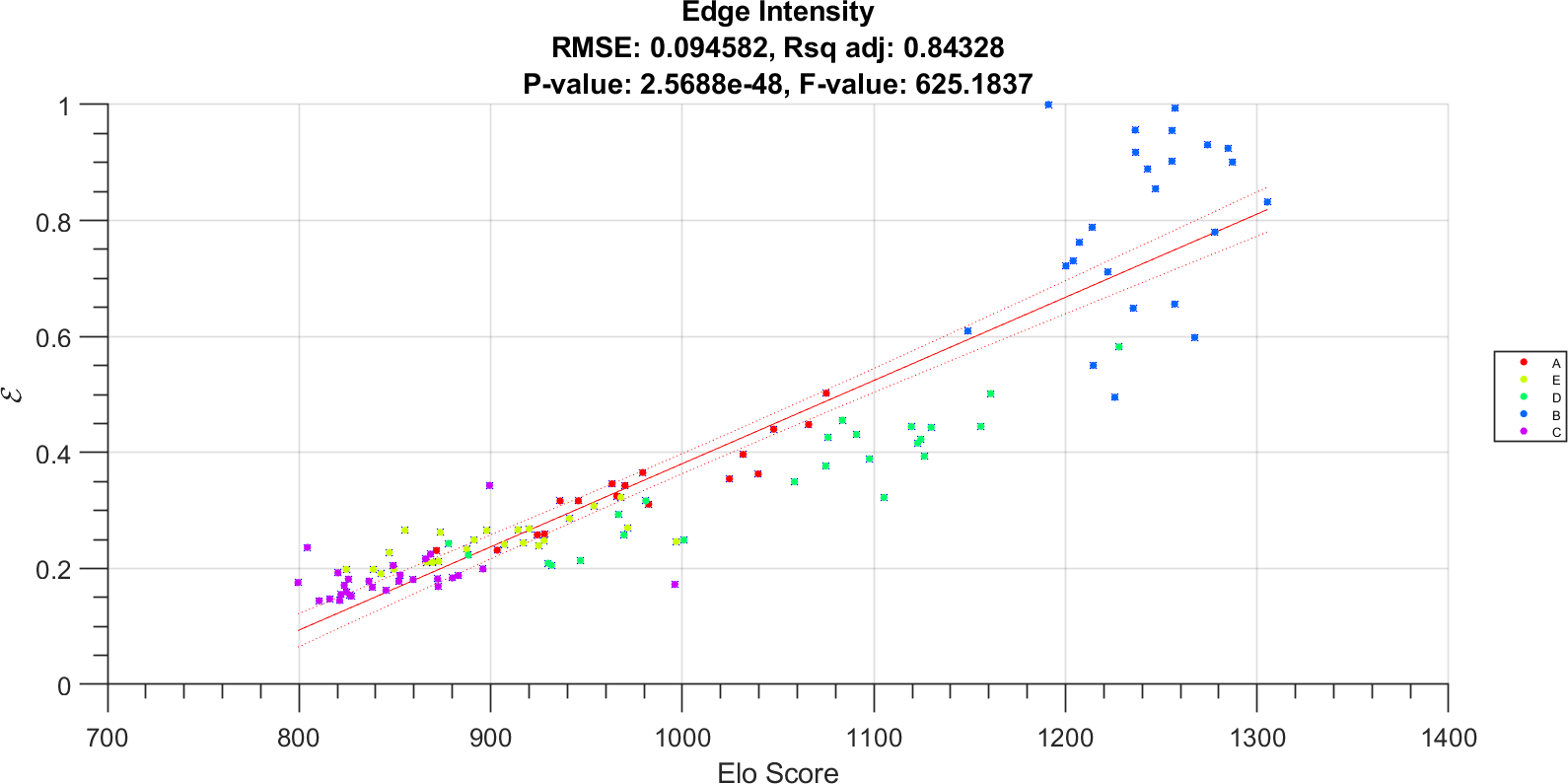}
  \caption{A plot of the edge intensity against the Elo score.}\label{fig:edgeIntensity}
\end{figure}

\begin{figure}[htbp]
  \centering
  \includegraphics[width=0.85\textwidth]{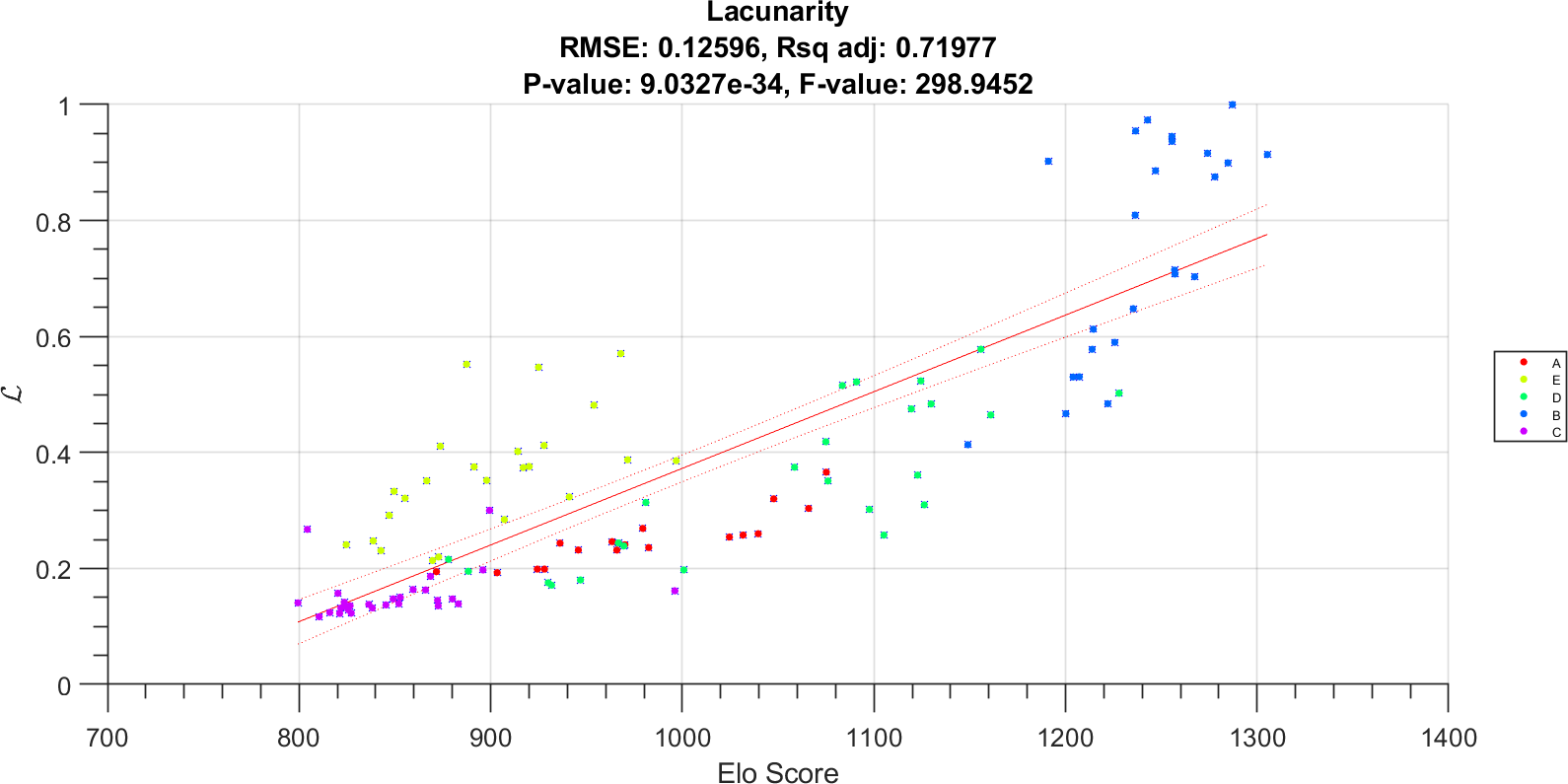}
  \caption{A plot of the lacunarity against the Elo score.}\label{fig:lacunarity}
\end{figure}

\begin{figure}[htbp]
  \centering
  \includegraphics[width=0.85\textwidth]{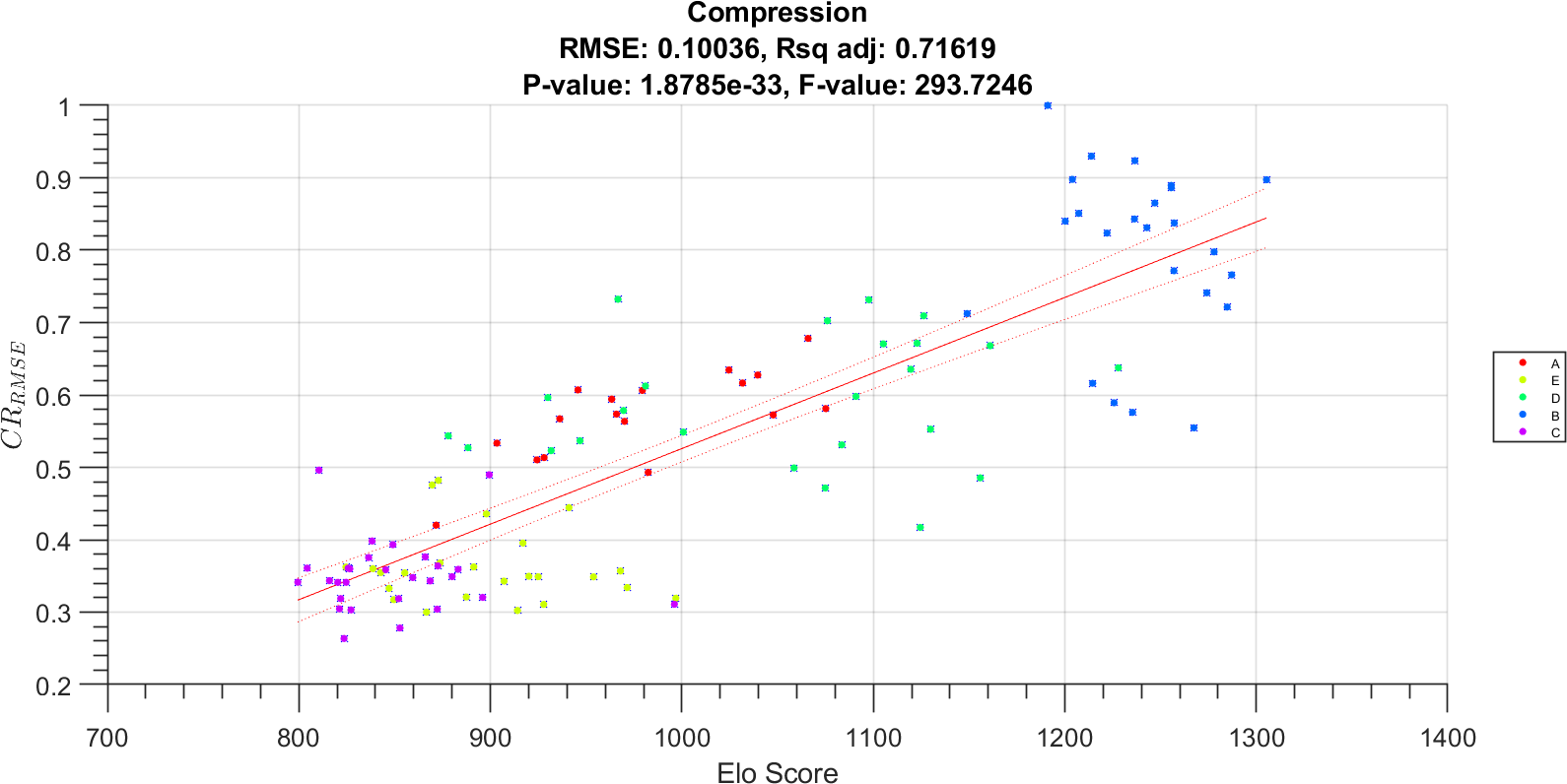}
  \caption{A plot of the compression ratio against the Elo score.}\label{fig:compression}
\end{figure}

\begin{figure}[htbp]
  \centering
  \includegraphics[width=0.85\textwidth]{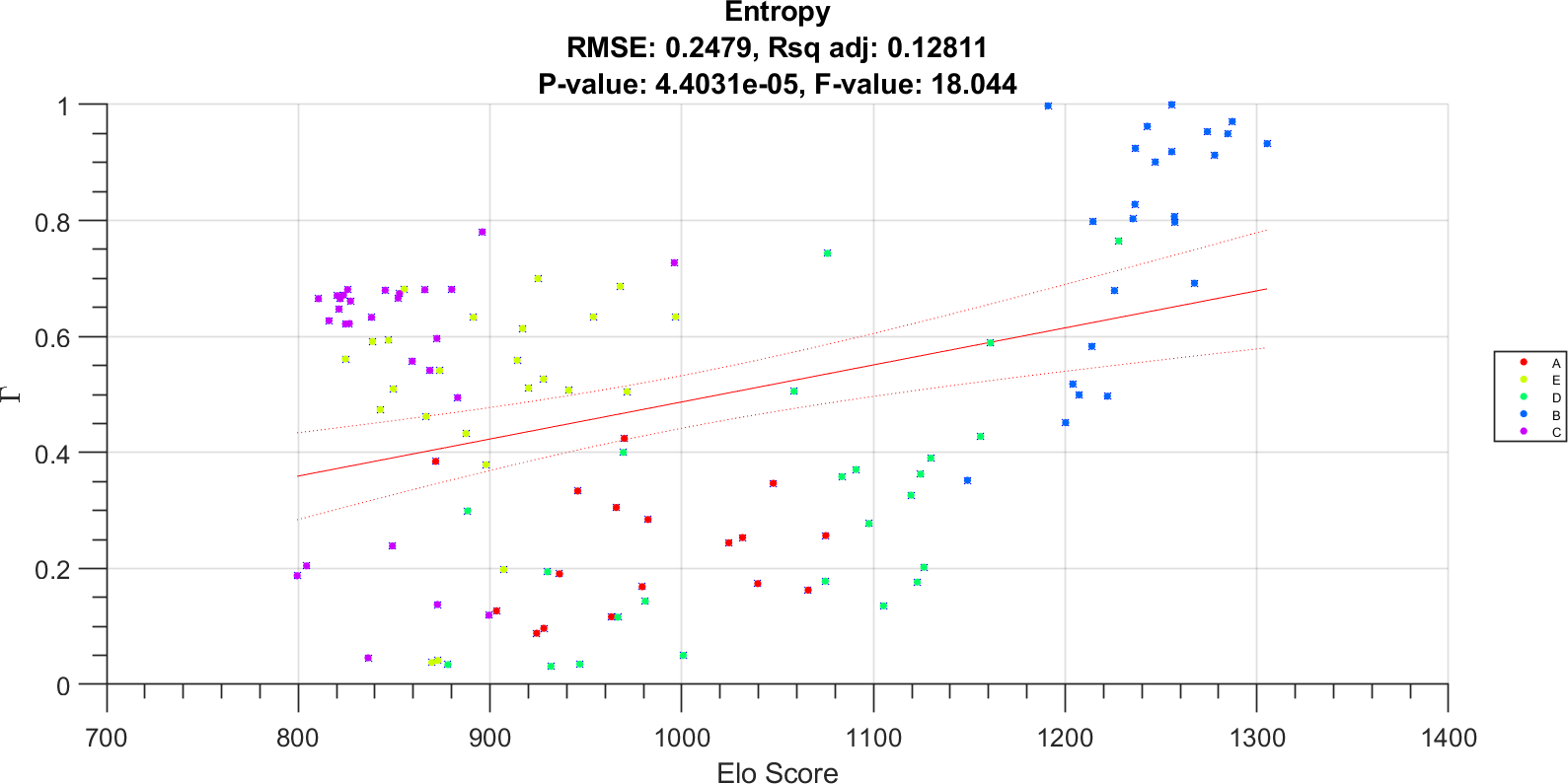}
  \caption{A plot of the entropy against the Elo score.}\label{fig:entropy}
\end{figure}

The image metrics were computed for each of the 117 images compared in the Elo analysis.  Linear regression was performed with the image metric as the dependent variable and the Elo score as the independent variable. The result of the linear regression is shown in Figures \ref{fig:edgeIntensity}, \ref{fig:lacunarity}, \ref{fig:compression}, and \ref{fig:entropy}.  Both the lacunarity and the edge intensity showed good correlation with the Elo score with R-squared values of 0.72 and 0.84.  Compression using the JPEG image format also showed a high degree of correlation with the complexity with an R-squared value of 0.72.  However, the Elo score using the entropy did not have a high degree of correlation with the Elo score with an R squared value of 0.13.

\section{Conclusions}
In this work, an image complexity ranking system was devised using the Elo rating system.  This enabled the rank ordering of a set of images in terms of their relative complexity as perceived by a set of operators.  Most operators had good agreement as to the complexity of the images with one SME having a slightly different ordering of the imagery. Large ripples, on the order of 1 m center to center, were perceived to have the highest degree of complexity whereas benign sand bottoms had lowest.

A set of common complexity metrics were computed from the image set and compared to the Elo scores.  It was found that measures of spatial variation had the highest degree of correlation with the human assessed complexity when compared to measures of total information content, with the measure of edge intensity having an R-squared value of approximately 0.84.

In the future, this test will be extended to a larger dataset to incorporate complexity from a large diversity of environments.  Further, there are many image complexity metrics that were not explored in this study.  For instance, the fractal dimension as well as anisotropy were not measured but have been shown to be important metrics when distinguishing image textures \cite{mandelbrot1982fractal}.  Therefore, in future testing more metrics will be incorporated and compared to the assessed image complexity.


\FloatBarrier

\printbibliography
\end{document}